\documentclass[sigconf]{acmart}
\renewcommand\footnotetextcopyrightpermission[1]{} 
\copyrightyear{2024}
\acmYear{2024}
\setcopyright{acmlicensed}\acmConference[ICCAD '24]{IEEE/ACM International Conference on Computer-Aided Design}{October 27--31, 2024}{New York, NY, USA}
\acmBooktitle{IEEE/ACM International Conference on Computer-Aided Design (ICCAD '24), October 27--31, 2024, New York, NY, USA}
\acmDOI{10.1145/3676536.3676756}
\acmISBN{979-8-4007-1077-3/24/10}

\graphicspath{{./images/}} 

\usepackage{multirow}
\usepackage{makecell}
\usepackage{subfig}
\usepackage{amsmath}

\begin{document}


\title{CFIRSTNET: Comprehensive Features for Static IR Drop Estimation with Neural Network}


\author{Yu-Tung Liu}
\authornote{Both authors contributed equally to this research.}
\email{tonyliu.ee09@nycu.edu.tw}
\orcid{0009-0003-8788-9104}
\affiliation{
  \institution{Department of Electronics and Electrical Engineering}
  \institution{National Yang Ming Chiao Tung University}
  \city{Hsinchu}
  \country{Taiwan}
}

\author{Yu-Hao Cheng}
\authornotemark[1]
\orcid{0009-0005-0946-4551}
\email{jason.ee09@nycu.edu.tw}
\affiliation{
  \institution{Department of Electronics and Electrical Engineering}
  \institution{National Yang Ming Chiao Tung University}
  \city{Hsinchu}
  \country{Taiwan}
}

\author{Shao-Yu Wu}
\orcid{0009-0007-6750-1415}
\email{sean31.ee09@nycu.edu.tw}
\affiliation{
  \institution{Department of Electronics and Electrical Engineering}
  \institution{National Yang Ming Chiao Tung University}
  \city{Hsinchu}
  \country{Taiwan}
}

\author{Hung-Ming Chen}
\email{hmchen@nycu.edu.tw}
\orcid{0000-0001-8173-3131}
\affiliation{
    \institution{Institute of Electronics}
    \institution{National Yang Ming Chiao Tung University}
    \city{Hsinchu}
    \country{Taiwan}
}

\renewcommand{\shortauthors}{Liu and Cheng, et al.}

\begin{abstract}
IR drop estimation is now considered a first-order metric due to the concern about reliability and performance in modern electronic products. Since traditional solution involves lengthy iteration and simulation flow, how to achieve fast yet accurate estimation has become an essential demand. In this work, with the help of modern AI acceleration techniques, we propose a comprehensive solution to combine both the advantages of image-based and netlist-based features in neural network framework and obtain high-quality IR drop prediction very effectively in modern designs. A customized convolutional neural network (CNN) is developed to extract PDN features and make static IR drop estimations. Trained and evaluated with the open-source dataset, experiment results show that we have obtained the best quality in the benchmark on the problem of IR drop estimation in ICCAD CAD Contest 2023, proving the effectiveness of this important design topic. 
\end{abstract}






\maketitle

\section{Introduction}
\label{sec:intro}
\begin{figure}[hbt!]
    \centering
    \includegraphics[width=\linewidth]{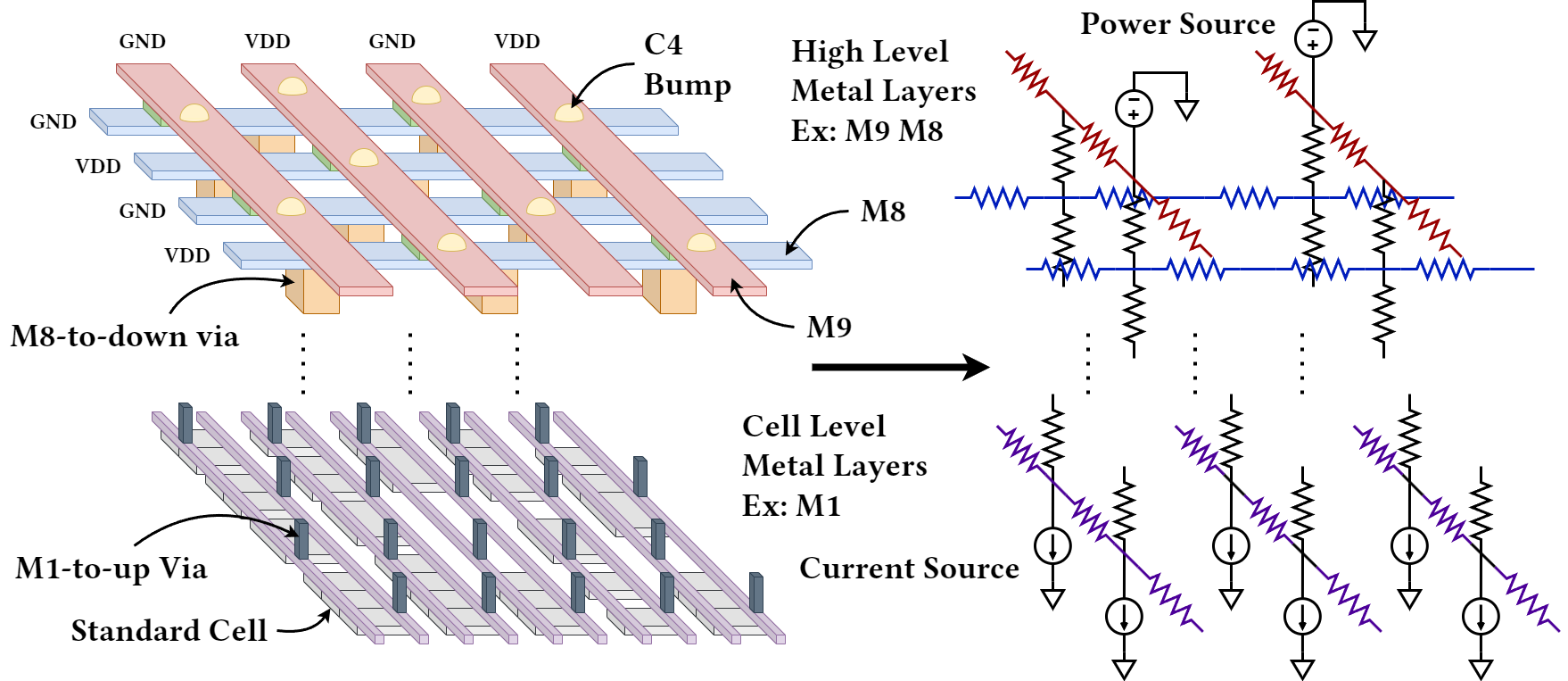}
    \caption{Illustration of a typical PDN structure: a resistive network between the power sources (C4 bumps) and the standard cells.}
    \label{fig: pdn}
\end{figure}

Analyzing the on-chip power delivery network (PDN) is critical to the modern integrated circuit (IC) design flow. Figure \ref{fig: pdn} is an illustration of a typical PDN. PDNs could be modeled as a network of resistance connecting the power pads (C4 bumps) and the standard cells. Static IR drop verification is an imperative step in PDN analysis. With the advance of technology, the IR drop issue occurs at the lower nodes leading to performance degradation. Moreover, excessive IR drop might also cause functional failure. Chip designers are supposed to have IR drop sign-off before the tape-out process and ensure to meet the IR drop constraint. If there is any IR drop violation, designers might need to perform the optimization step such as Engineer Change Order (ECO) to reorganize the circuit and attempt to resolve the problem. 

However, during every iteration of ECO, EDA tools will assess the circuit repeatedly, i.e. the worst-case voltage drop of the circuit will be calculated. The conventional method to retrieve the voltage of every node was solving the linear equation (\ref{eq: ir-drop}) where $G$ was a conductance matrix, $V$ the target voltage vector to be solved, and $J$ the vector of independent current sources. The number of nodes surged as the amount of cells used in modern design increased tremendously. The considerable computational time of the conventional nodal analysis method was so expensive that it became a costly overhead. Therefore, several methods were proposed to trade accuracy for speed and replace the traditional solution with a faster and more accurate IR drop estimation. 

\begin{equation}
\label{eq: ir-drop}
GV = J
\end{equation}

While previous works have proposed various approaches to trade accuracy with speed, they all faced limitations. Non-machine learning methods were swamped by complex circuits with hefty amounts of nodes. On the other hand, image-based or netlist-based machine learning (ML) methods focused on certain circuit features solely. We propose CFIRSTNET, a CNN approach along with comprehensive feature extraction to speed up the static IR drop estimation problem with minimal accuracy trade-off. In addition to the provided image-based features, netlist-based inputs are analyzed with KVL, KCL, and superposition in the feature extraction stage to model the electrical features of the PDN. The custom CNN model takes image-based features and netlist-based features for input and generates an IR drop prediction map with reasonable scaling, providing more accurate and more efficient estimations. Tested by the ICCAD CAD Contest 2023 open-source benchmark, CFIRSTNET mitigates the overhead of expensive computational resources and provides lower estimation errors in comparison to prior arts. This work could be further implemented in modern design flow and relieve the pain of heavy computational expenses.

The main contributions of this paper are listed as follows: 

\begin{itemize}
  \item CFIRSTNET harnesses the benefit of both image-based and netlist-based inputs and introduces a customized CNN model to perform static IR drop estimation. To the best of our knowledge, this is the first work regarding this combination as CNN model inputs that preserve circuit features while ensuring efficiency. 
  \item We propose the Hypothetical IR Drop Distillation that analyzes the power grid from the SPICE-based data in linear time and provides more comprehensive features for the customized CNN simultaneously. 
  Our approach is independent of fabrication technologies, circuit designs (with different current source and voltage source distributions), and PDN structures. 
  \item The proposed model is scalable due to the properties of CNNs. CFIRSTNET has been tested with various chip sizes and has exhibited strong estimation results. 
  The performance of CFIRSTNET is not affected by macros or irregular PDN arrangement. Compared to 1st place winner of ICCAD CAD Contest 2023, CFIRSTNET reduces average error by 60\%, max error by 66\%, increases F1 score by 59\%, and achieves a 21.8x speedup. 
  
\end{itemize}

The remainder of this paper is organized as follows. Preliminaries to this problem are discussed in section \ref{sec: prelim}. The details of CFIRSTNET will be elaborated in section \ref{sec: comp feature} and \ref{sec: model}. Section \ref{sec: exp} and \ref{sec: result} present the experimental setup and results. Finally, Section \ref{sec: conclusion} concludes the paper and discusses the possibilities of exploring future works.

\section{Preliminaries}
\label{sec: prelim}

Previously proposed methods introduced non-ML methods by modeling the whole power grid and finding the solution to Equation (\ref{eq: ir-drop}). Panda et al.\cite{MNA} provided a modeling way of resistance-inductance-capacitance (RLC) circuits. Moreover,  Panda et al.\cite{MNA} and Zhao et al.\cite{hierarchical2002} ignored the influence of the inductance in lower layers due to the small number compared to resistance and capacitance. This simplicity made some matrices symmetric and positive definite, which enabled Cholesky factorization to speed up the works.

When it comes to tens or even hundreds of millions of nodes in the circuit, flattened circuit models require tremendous memory capacity and computational resources. Zhao et al.\cite{hierarchical2002} proposed a hierarchy method to lower the memory cost, and speed up the circuit simulation. They generated micromodels for local grids and simulated the whole network. Kozhaya et al.\cite{kozhaya2002multigrid} assumed that the sink resistance is larger than the grid resistance, which bridged the gap of the voltage drop between nearby grids. Hence, they used a more coarse structure to process the PDN and accelerate the framework by mapping out the solution with interoperation. The methods mentioned above traded off the accuracy to the speed. Other works such as Qian et al.\cite{randomwalkqian2005power} took the hierarchical random-walk approach. However, the run time of these simulated methods was still prone to the increased node number.

Recently, many ML methods have cast light on the EDA problems, few of which explored the static IR drop estimation problem. Compared with the traditional simulation-based methods, the runtime of ML methods is not affected by the number of circuit nodes. Large circuits could be processed and the predictions could be made in an unprecedentedly short amount of time. The ML models could be classified into two categories according to the selection of input features. Some took the circuit netlist as input and decision-tree models were introduced\cite{chen2022vector}\cite{fang2018machine}\cite{IncPIRD_ho2019incpird}\cite{pao2020xgbir}. Others formulated the problem as an IR drop map prediction model\cite{chhabria2021thermal}\cite{fang2018machine}\cite{xie2020powernet}\cite{zhou2020gridnet}. Moreover, previous ML-based IR drop prediction methods focused on static and dynamic IR drop estimation\cite{chen2022vector}\cite{chhabria2021mavirec}\cite{fang2018machine}. The two categories were based on different problem settings that dynamic IR drop estimation captures the peak transient current values based on switching activities, which is not available in open-source datasets.

Previous ML works could be separated by the use of input data. Fang et al.\cite{fang2018machine} involved CNN to perform feature extraction and make predictions with XGBoost\cite{chen2016xgboost}. Ho et al.\cite{IncPIRD_ho2019incpird} assessed the circuit netlist and the technology libraries and extracted more features with superposition and partition. However, the feature extraction algorithms took great time complexity. On the other hand, Chhabria et al.\cite{chhabria2021thermal} took advantage of the CNNs with image-based inputs. Unfortunately, the resolution of the proposed image-based data for the experiment was too coarse to indicate the actual IR drop values, which will be discussed in section \ref{sec: exp}. 


To unleash the possibilities of ML methods in static IR drop estimation problems, substantial training data is necessary. Though the approaches mentioned above have achieved the best performance with their restricted evaluation data points, open-source benchmark data is crucial to make a direct comparison among all methods. However, publicly available real PDN circuit netlists are limited due to the confidentiality agreement and some constraints from EDA vendors. Kadagala et al.\cite{kadagala20232023} and Chhabria et al.\cite{chhabria2021began} noticed this lack of an open-source dataset and proposed the dataset consists of real and fake circuits. Chhabria et al.\cite{chhabria2021began} address this issue with Generative Adversarial Networks (GANs) connected with OpeNPDN\cite{opeNPDN} to generate current maps and power girds with netlist form and release thousands of generated data in different technologies. In our work, we collect PDN circuit data from \cite{kadagala20232023} and \cite{chhabria2021began} under the open-source NanGate 45nm technology\cite{ajayi2019openroad} for our experiments. 

\noindent \textbf{Problem Formulation}

\noindent \textbf{Given: } Circuit netlist in SPICE format, current map, effective distance to voltage source map, and PDN density map.

\noindent \textbf{Train: } Extract comprehensive circuit features to train a custom CNN model to make static IR drop estimations.

\noindent \textbf{Output: } Static IR drop estimation map that matches the size of the ground truth.

\begin{figure*}[htb!]
    \centering
    \includegraphics[width=0.9\textwidth]{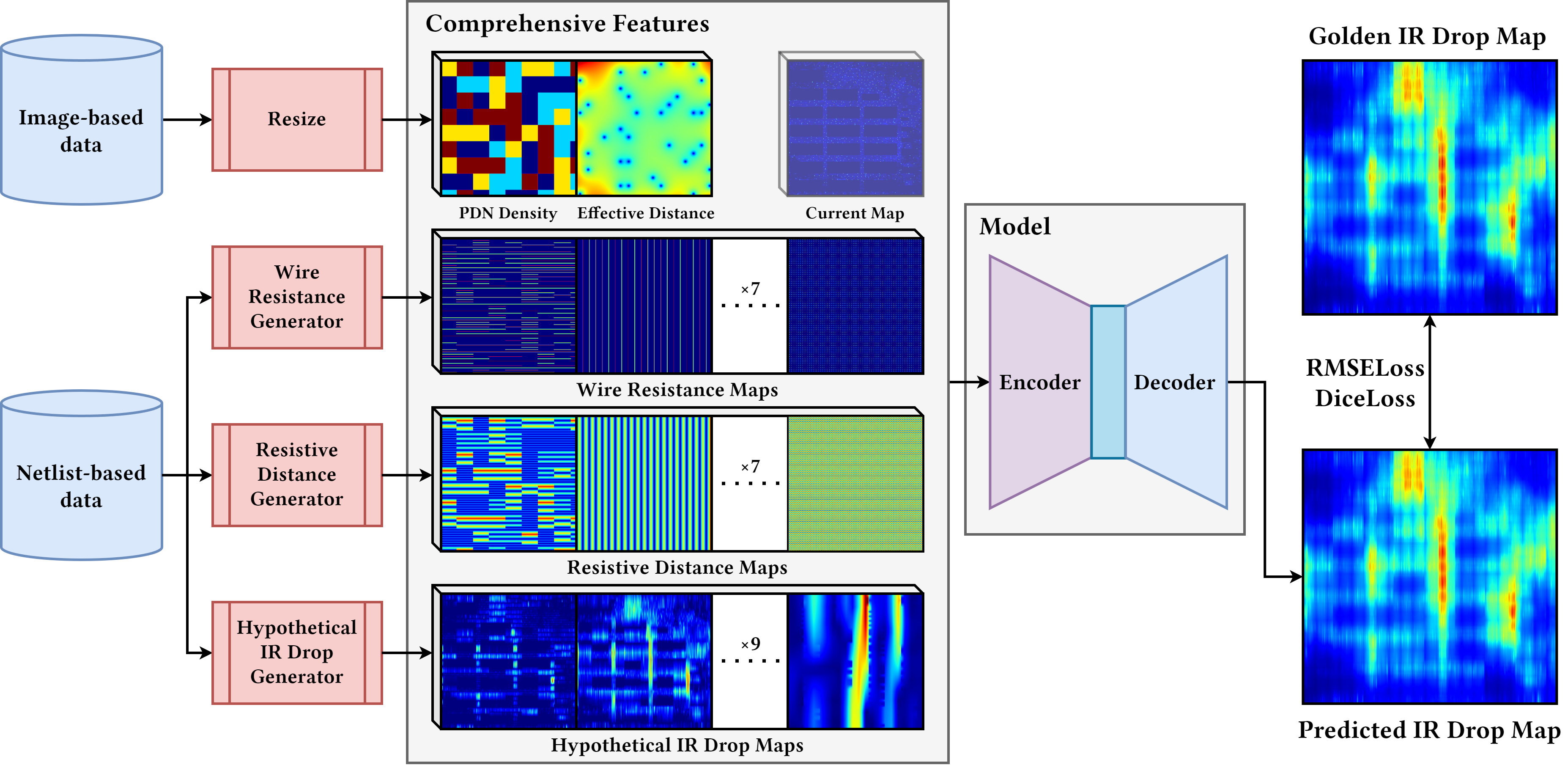}
    \caption{Extracted features and the CFIRSTNET prediction flow: Along with the provided image-based data, CFIRSTNET extracts augmented features from the netlist-based data and processed by the custom CNN.}
    \label{fig: flow}
\end{figure*}

\section{Comprehensive Features}

\label{sec: comp feature}

CFIRSTNET combines the best of the netlist-based and image-based features. An overview of the CFIRSTNET prediction flow and the roles of the feature maps is illustrated in Figure \ref{fig: flow}. The current map, PDN density map, and effective distance to voltage source map are provided in the open-source benchmark\cite{kadagala20232023}. Additionally, we propose augmented feature maps such as the wire resistance maps, the resistive distance maps, and the hypothetical IR drop maps extracted from the SPICE netlist to provide more information to our custom CNN. The comprehensive feature maps will be processed by the encoder-decoder CNN. The encoder is comprised of a modified ConvNeXtV2\cite{woo2023convnextv2} encoder block, while the decoder includes a custom feature pyramid network (FPN) and an IR drop reconstruction module. The output IR drop prediction map will be generated from CFIRSTNET and evaluated by the open-source benchmark.

The additional feature maps represent the electrical and geometric features of the power delivery network. They are collected as feature maps that could be concatenated with the provided image-based features as the model inputs. The following sections will elaborate on the comprehensive model input features.

\subsection{Provided Image-Based Maps}
The benchmark dataset\cite{kadagala20232023} used in this work provides three image-based data: a current map, an effective distance to voltage source map, and a PDN density map. The current map provides the power-consuming hotspots and the effective distance to voltage source map indicates the distance to the power sources. Notably, the benchmark dataset in \cite{kadagala20232023} is comprised of regular and irregular PDN designs. Therefore, the PDN density map offers the wire density information of every part of the circuit. 

\subsection{Wire Resistance Maps}
The challenge of estimating IR drop in ICs hinges on understanding the interplay between the power source, power-consuming cells, and the underlying power grid. In essence, the distribution of the power grid and the actual power consumption significantly impact the IR drop. Previous research\cite{chhabria2021thermal}\cite{fang2018machine}\cite{xie2020powernet}\cite{zhou2020gridnet} has leveraged the localization and feature extraction capabilities of CNNs to pinpoint the power source, delineate the power grid, and quantify current consumption. These efforts have primarily focused on 2D feature maps for image-to-image IR drop prediction networks.

However, when dealing with complex ICs, which are inherently 3D structures with multiple metal layers connecting transistors and I/O pins, relying solely on 2D feature maps may overlook crucial details. For instance, the intricate arrangement of different metal layers and the vias (connections) between them demands a more nuanced approach.

CFIRSTNET models the power grid with multiple-layer resistance maps derived from the spice netlist. By concatenating these maps with other relevant features, we aim to capture finer-scale information such as the precise distribution of the power grid and the positions of vias for our subsequent CNN model. This approach promises a more accurate representation of the complex 3D circuitry, enhancing our ability to predict IR drop effectively.

\subsection{Resistive Distance Maps}
When the distance from a cell, a power sink, to the power source, is considerable, it might require longer metals of different layers to connect the cell to the power source, and longer metals indicate larger resistance, which will lead to a larger IR drop. To model this property, we calculate the shortest Manhattan distance from every part of the power delivery network to every metal wire and via point. Furthermore, we collect this information and concatenate it with other feature maps.

\subsection{Hypothetical IR Drop Distillation}
The IR drop of each power node is greatly affected by nearby nodes and power pads due to the local characteristics of PDNs. Traditional IR drop solvers\cite{kozhaya2002multigrid}\cite{hierarchical2002} often partition the whole PDN into several smaller circuit segments to reduce complexity. Meanwhile, an increased error of IR drop value might be found after partitioning since the PDN is not considered as a whole. To mitigate this issue, previous work \cite{zhong2005fast} introduced iterative methods, which will impact the overall runtime considerably.

CFIRSTNET introduces an innovative methodology that partitions the PDN into smaller segments with some additional assumptions. The main idea is that these hypothetical IR drops could be hints for the customized CNN model to make more precise estimations. The hypothetical IR drop values of these circuit segments can be retrieved in linear time complexity. Furthermore, these hypothetical IR drops will be part of our model input features.

\subsubsection{Localized IR Drop}
First of all, PDNs consist of multiple metal layers connected with vias. Each layer could be viewed as a plane with plentiful metal stripes. Due to 
preferred metal wire direction, there is no interconnection between each stripe. Thus, we narrow our scope to a single metal stripe with wire resistances, current sources, and voltage sources. 

The computation of localized IR drops hinges on the assumption that vias connecting to the upper layer function as voltage sources and each has the value of VDD, while those linking to the lower layer serve as current sources. The computation of the localized IR drops and current of each instance starts from the lowermost layer. The retrieved current to the upper vias will be the current sources in the layer above. The current source at node $n$ is defined as $I_{n}$, and the voltage as $V_{n}$. Node $n$ is the $n_{th}$ node on the N-node stripe, and the wire resistance between node $m$ and node $n$ is represented by $r_{m,n}$.

\begin{figure}[hbt!]
    \includegraphics[width=\linewidth]{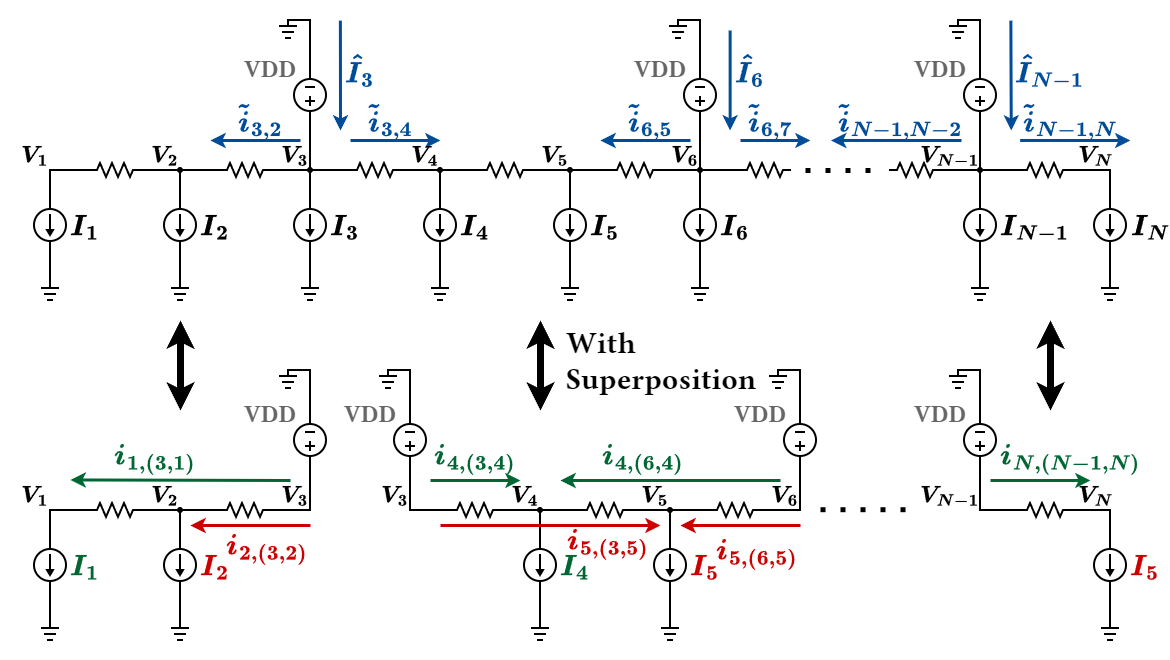}
    \caption{Metal stripe and circuit segments.}
    \label{fig: wire}
\end{figure}

Furthermore, we can analyze the circuit separately by superposition. As shown in Figure \ref{fig: wire}, each smaller circuit segment contains one or two voltage sources.

The equivalent resistance $R_{n}$ between node $n$ and voltage sources could be obtained by Equation (\ref{eq: parallel}), where $P$ is the set of voltage sources in the circuit segment. 

\begin{equation}
    \label{eq: parallel}
    R_{n} = \dfrac{1}{\sum_{m \in P}\dfrac{1}{r_{m,n}}}
\end{equation}

With superposition, each current source $I_{n}$ could be evaluated independently as well. The current $i_{n,(k,l)}$ caused by $I_{n}$ from node $k$ to $l$ could be calculated by Equation (\ref{eq: current}). 

\begin{equation}
    \label{eq: current}
    i_{n,(k,l)}=
    \begin{cases}
        I_{n}\dfrac{R_{n}}{r_{m,n}}, & \text{if  \(k\) and  \(l\) are between  \(m\) and  \(n\)}\\
        0, & \text{otherwise} \\
    \end{cases}
\end{equation}

The current $\tilde{i}_{k,l}$ response with all current sources in effect between each node could be sum algebraically in (\ref{eq: sum_current}).

\begin{equation}
    \label{eq: sum_current}
    \tilde{i}_{k,l}=\sum_{n=1}^{N}i_{n,(k,l)}
\end{equation}

According to KVL, the voltage drop from node $n$ to $n+1$ is represented in Equation (\ref{eq: voltage_i}).

\begin{equation}
    \label{eq: voltage_i}
    V_{n+1}=V_{n} - \tilde{i}_{n,n+1}r_{n,n+1}
\end{equation}

Equation (\ref{eq: voltage_i}) can be reformulated as Equation (\ref{eq: v_i_recur}) after applied recursively, where $p \in P$ and $IR_{pn} = \sum_{k=p}^{n-1}\tilde{i}_{k,k+1}r_{k,k+1}$ is the localized IR drop voltage. Consequently, the voltage response on node $n$ is presented in Equation (\ref{eq: v_i_recur}). 

\begin{equation}
    \label{eq: v_i_recur}
    V_{n} = V_{p} - IR_{pn}
\end{equation}

The current $\hat{I}_{n}$ on voltage source $n$, could be obtained by Equation (\ref{eq: i_pi}), where $I_{n}$ is the current source on node $n$.

\begin{equation}
    \label{eq: i_pi}
    \hat{I}_{n}=I_{n} + \tilde{i}_{n,n-1} + \tilde{i}_{n,n+1}
\end{equation}

Eventually, the IR drop $IR_{via,n}$ across via $n$ is expressed in Equation (\ref{eq: via_drop}), and the IR drop on every node is acquired. We could pass these properties to the next stage and retrieve the overall circuit feature. 

\begin{equation}
    \label{eq: via_drop}
    IR_{via,n}=\hat{I}_{n}R_{via,n}
\end{equation}

\begin{figure*}[t!]
    \centering
    \includegraphics[width=0.9\textwidth]{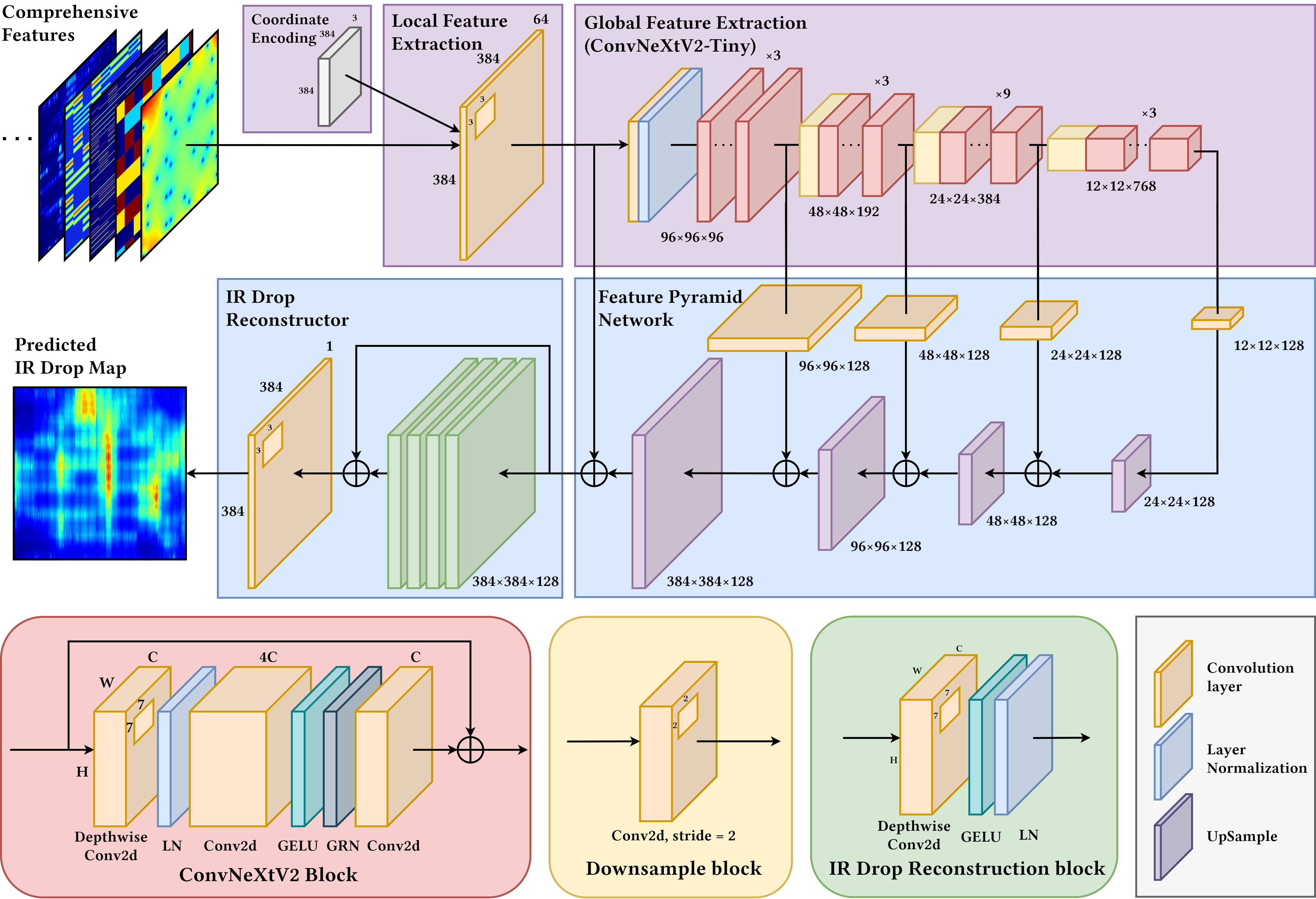}
    \caption{The detailed illustration of custom CNN model architecture in Figure \ref{fig: flow}. CFIRSTNET extracts the latent representation of the PDN from the model input features in the encoder stage (purple), aggregates features of various resolutions with the FPN (blue), and generates a high-resolution IR drop estimation map with the custom reconstructor.}
    \label{fig: architecture}
\end{figure*}

\subsubsection{Hypothetical IR Drop}

Upon analyzing each metal layer, we perceive the PDN as a cohesive unit. Our primary goal is to determine the IR drop specifically on the cell level. Having already obtained the localized IR drop values for each node of the circuit individually, we aim to figure out how these cumulative IR drops from each metal layer impact the base layer. Starting from the top layer, consider the IR drop at each node associated with a via. These IR drops effectively act as voltage sources for the subsequent layer.  According to KCL, the aggregate current at node $i$ of metal stripe $s$ in layer $m$ must be zero (as shown in Equation (\ref{eq: kirchhoff})), where $V_{m,s,i}$ represents the voltage on node $i$ of metal stripe $s$ in layer $m$ and $R_{m,s,i,k}$ the resistance between node $i$ and $k$ on stripe $s$ of metal $m$. 

\begin{equation}
    \label{eq: kirchhoff}
    \sum_{k \in N_p}\dfrac{V_{m,s,i}-V_{m,s,k}}{R_{m,s,i,k}} = 0
\end{equation}

Consequently, Equation (\ref{eq: kirchhoff}) could be elaborated as Equation (\ref{eq: vmi}) by means of voltage division. 

\begin{equation}
    \label{eq: vmi}
    V_{m,s,i} = \sum_{k \in N_p}V_{m,s,k}\dfrac{\dfrac{1}{R_{m,s,i,k}}}{\sum_{j \in N_p}\dfrac{1}{R_{m,s,i,j}}}
\end{equation}\textbf{}

Eventually, we plot the hypothetical IR drop, which is the effect of each metal layer on the cell level, separately. Applying the one-dimensional interpolation on two axes could foster the hypothetical IR drop map. As a result, the generated feature could be transferred to the intrinsic model input.

\section{Model Architecture}
\label{sec: model}

CFIRSTNET includes a CNN with an encoder-decoder architecture, commonly used in computer vision tasks. Unlike traditional symmetric CNNs, CFIRSTNET features a customized encoder and decoder, specifically designed for the problem at hand. The tailored encoder extracts both shallow local features and deep global features. Moreover, the amalgamative decoder merges features from multiple resolutions, combining global features with local features. Figure \ref{fig: architecture} is a visual representation of our model. CFIRSTNET extracts the latent representation of the PDN from the model input features in the encoder stage (purple), aggregates features of various resolutions with the FPN (blue)\cite{FPN}, and generates a high-resolution IR drop estimation map with the custom reconstructor. The unique design of CFIRSTNET ensures efficient feature extraction and resolution adaptation.

\subsection{ConvNeXtV2 Inherited Encoder}

We aim to derive a hierarchical feature maps that captures the global representation of individual IR drop values across various scales. Unlike image classification tasks, which often benefit from extensive annotated datasets, the scarcity of large annotated datasets for IR drop poses a challenge. To address this, we opt for ConvNeXtV2\cite{woo2023convnextv2}, an enhanced variant of the original ConvNeXt architecture\cite{liu2022convnet}. ConvNeXtV2 is specifically chosen for its ability to extract 2-D spatial information effectively. Its inductive bias towards convolution helps prevent overfitting, and its large receptive field ensures robust global representation. In the context of static IR drop estimation, it is essential to analyze the circuit using small segments to capture fine-grained local features. Simultaneously, we must consider the entire circuit hierarchy through global representations. The CFIRSTNET encoder comprises three major working blocks: coordinate encoding, local feature extraction, and global feature extraction.

\subsubsection{Coordinate Encoding}

A typical CNN exhibits translation invariance, meaning that its output remains consistent regardless of the spatial position of input objects. However, in the context of PDN synthesis patterns used in EDA tools, the issue of IR drop becomes highly relevant to the placement of circuit components. To address this issue, we introduce a coordinate encoding layer\cite{coordconv} in CFIRSTNET prior to the feature extraction stage. This layer encodes translation information related to the input data and concatenates it as augmented input channels. Consequently, CFIRSTNET effectively maps the spatial position of each pixel.

\subsubsection{Local Feature Extraction}
We employ a standard convolutional layer with a 3x3 kernel size to extract local features. This layer provides CFIRSTNET with the initial latent representation of the input circuit.

\subsubsection{Global Feature Extraction}

ConvNeXt\cite{liu2022convnet} is a modernized ConvNet for image classification and segmentation. It combines large receptive fields with scalability, drawing inspiration from Vision and Swin Transformers\cite{2020vit}\cite{liu2021swin}. At the macro design level, ConvNeXt follows Swin Transformer's stage ratio and patchify stem design. For block design, it increases the kernel size to 7x7 (compared to original ResNet\cite{resnet}) while using depthwise convolution to maintain FLOPS. The inverted bottleneck design from Transformers transforms the block design into a depthwise convolution layer followed by a multilayer perceptron (MLP) with an expansion ratio of 4. At the micro design level, ConvNeXt employs GELU activation and Layer Normalization (LN). ConvNeXtV2\cite{woo2023convnextv2} enhances this by adding a Global Response Normalization (GRN) layer after the GELU activation function and utilizing Fully Convolutional Masked Autoencoder (FCMAE) pre-training for improved performance compared to supervised pre-training.

\subsection{Feature Fusion Decoder}
To fuse features across different hierarchies and capture fine-grained details of IR drop values, we design our decoder in two parts: a FPN\cite{FPN} and an IR drop reconstruction module. The FPN enables feature fusion at various levels, while the IR drop reconstruction module reconstructs the IR drop value using the fused features.

\subsubsection{Feature Pyramid Network}

The FPN in CFIRSTNET leverages multi-level features obtained from the backbone. These features are computed at several scales with a scaling step of 2. The bottom-up pathway computes a feature hierarchy, resulting in feature maps at different levels. FPN then transforms features from each level into a unified latent space using convolution and upsampling. This process ensures that latent features maintain their original sizing. The top-down pathway enhances spatially coarser, but semantically stronger, feature maps from higher pyramid levels. The result is a set of proportionally sized feature maps at multiple levels. In our custom FPN design, we input both local features and fine-grained and global features. FPN acts as a generic solution, independent of the backbone convolutional architectures, and provides fused features for IR drop map reconstruction.

\subsubsection{IR Drop Reconstruction}
We develop the IR drop reconstruction module by employing a series of four cascaded depth-wise convolutions with a 7x7 kernel size, complemented by LN. The activation function layers utilize GELU, and we establish a residual connection from the base layer of the FPN. Ultimately, a 3x3 convolutional layer produces the final IR drop prediction map. This module design maintains the crucial fine-grained feature encoding necessary for generating high-resolution IR drop maps.

\begin{table}[ht]
\centering
\caption{Detail Information of PDNs for Testing.}
\smallskip
\label{tab: circuit}
\resizebox{0.89\columnwidth}{!}{
\begin{tabular}{|c|c|c|c|}
\hline
\multirow{2}{*}{Instance} & \multirow{2}{*}{\begin{tabular}[c]{@{}c@{}}With macros\end{tabular}} & \multirow{2}{*}{\begin{tabular}[c]{@{}c@{}}Chip size  (mm$^{2}$)\end{tabular}} & \multirow{2}{*}{\begin{tabular}[c]{@{}c@{}}PDN type\end{tabular}} \\
 &  &  &  \\ \hline
testcase7 & \multirow{2}{*}{O} & \multirow{2}{*}{0.361} & irregular \\ \cline{1-1} \cline{4-4} 
testcase8 &  &  & regular \\ \hline
testcase9 & \multirow{2}{*}{O} & \multirow{2}{*}{0.697} & irregular \\ \cline{1-1} \cline{4-4} 
testcase10 &  &  & regular \\ \hline
testcase13 & \multirow{2}{*}{X} & \multirow{2}{*}{0.066} & irregular \\ \cline{1-1} \cline{4-4} 
testcase14 &  &  & regular \\ \hline
testcase15 & \multirow{2}{*}{X} & \multirow{2}{*}{0.239} & irregular \\ \cline{1-1} \cline{4-4} 
testcase16 &  &  & regular \\ \hline
testcase19 & \multirow{2}{*}{O} & \multirow{2}{*}{0.757} & irregular \\ \cline{1-1} \cline{4-4} 
testcase20 &  &  & regular \\ \hline
\end{tabular}
}
\end{table}

\begin{figure}[h!]
    \includegraphics[width=\linewidth]{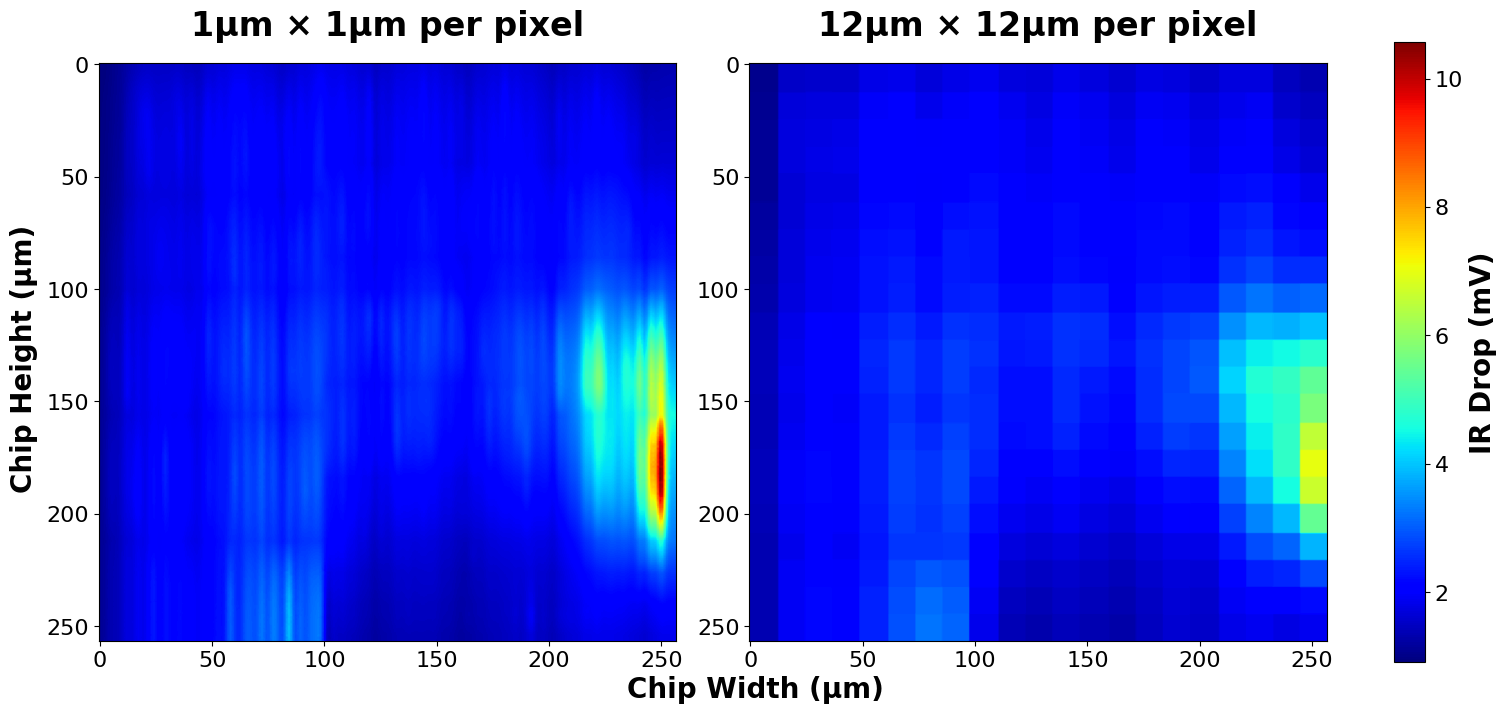}
    \caption{Comparison between IR drop maps under different resolutions. A lower resolution (12$\mu$m x 12$\mu$m) can not represent the actual IR drop values.}
    \label{fig: resolution}
\end{figure}

\begin{table*}[ht!]
\centering
\caption{Overall performance for IR drop estimation.}
\smallskip
\label{tab: result-table}
\resizebox{1\textwidth}{!}{

\begin{tabular}{|c|cccc|cccc|cccc|ccccc|}
\hline
\multirow{2}{*}{} & \multicolumn{4}{c|}{\multirow{2}{*}{HSPICE}} & \multicolumn{4}{c|}{\multirow{2}{*}{IREDGe}} & \multicolumn{4}{c|}{\multirow{2}{*}{1st Place of CAD Contest}} & \multicolumn{5}{c|}{\multirow{2}{*}{CFIRSTNET}} \\
 & \multicolumn{4}{c|}{} & \multicolumn{4}{c|}{} & \multicolumn{4}{c|}{} & \multicolumn{5}{c|}{} \\ \hline
Instance & \multicolumn{1}{c|}{\begin{tabular}[c]{@{}c@{}}$IR_{avg}$\\ (mV)\end{tabular}} & \multicolumn{1}{c|}{\begin{tabular}[c]{@{}c@{}}$IR_{max}$\\ (mV)\end{tabular}} & \multicolumn{1}{c|}{\begin{tabular}[c]{@{}c@{}}Hotspot\\ (\%)\end{tabular}} & \begin{tabular}[c]{@{}c@{}}Runtime\\ (sec)\end{tabular} & \multicolumn{1}{c|}{\begin{tabular}[c]{@{}c@{}}$e_{avg}$\\ (mV)\end{tabular}} & \multicolumn{1}{c|}{\begin{tabular}[c]{@{}c@{}}$e_{max}$\\ (mV)\end{tabular}} & \multicolumn{1}{c|}{F1 Score} & \begin{tabular}[c]{@{}c@{}}runtime\\ (sec)\end{tabular} & \multicolumn{1}{c|}{\begin{tabular}[c]{@{}c@{}}$e_{avg}$\\ (mV)\end{tabular}} & \multicolumn{1}{c|}{\begin{tabular}[c]{@{}c@{}}$e_{max}$\\ (mV)\end{tabular}} & \multicolumn{1}{c|}{F1 Score} & \begin{tabular}[c]{@{}c@{}}Runtime\\ (sec)\end{tabular} & \multicolumn{1}{c|}{\begin{tabular}[c]{@{}c@{}}$e_{avg}$\\ (mV)\end{tabular}} & \multicolumn{1}{c|}{\begin{tabular}[c]{@{}c@{}}$e_{max}$\\ (mV)\end{tabular}} & \multicolumn{1}{c|}{F1 Score} & \multicolumn{1}{c|}{\begin{tabular}[c]{@{}c@{}}Runtime\\ (sec)\end{tabular}} & \begin{tabular}[c]{@{}c@{}}Speedup\\ (vs HSPICE)\end{tabular} \\ \hline
testcase7 & \multicolumn{1}{c|}{1.0439} & \multicolumn{1}{c|}{4.3045} & \multicolumn{1}{c|}{0.378} & 8.94 & \multicolumn{1}{c|}{0.6218} & \multicolumn{1}{c|}{1.2305} & \multicolumn{1}{c|}{0.142} & \textbf{0.150} & \multicolumn{1}{c|}{0.0656} & \multicolumn{1}{c|}{1.2115} & \multicolumn{1}{c|}{0.783} & 7.996 & \multicolumn{1}{c|}{\textbf{0.0177}} & \multicolumn{1}{c|}{\textbf{0.3458}} & \multicolumn{1}{c|}{\textbf{0.923}} & \multicolumn{1}{c|}{0.366} & 24.43x \\ \hline
testcase8 & \multicolumn{1}{c|}{1.5540} & \multicolumn{1}{c|}{4.8994} & \multicolumn{1}{c|}{0.535} & 8.86 & \multicolumn{1}{c|}{0.3845} & \multicolumn{1}{c|}{2.2659} & \multicolumn{1}{c|}{0.419} & \textbf{0.149} & \multicolumn{1}{c|}{0.0815} & \multicolumn{1}{c|}{1.0416} & \multicolumn{1}{c|}{0.816} & 8.396 & \multicolumn{1}{c|}{\textbf{0.0257}} & \multicolumn{1}{c|}{\textbf{0.5214}} & \multicolumn{1}{c|}{\textbf{0.916}} & \multicolumn{1}{c|}{0.367} & 24.14x \\ \hline
testcase9 & \multicolumn{1}{c|}{1.1811} & \multicolumn{1}{c|}{3.7932} & \multicolumn{1}{c|}{0.034} & 21.07 & \multicolumn{1}{c|}{0.4538} & \multicolumn{1}{c|}{1.2780} & \multicolumn{1}{c|}{0} & \textbf{0.264} & \multicolumn{1}{c|}{0.0406} & \multicolumn{1}{c|}{0.8755} & \multicolumn{1}{c|}{\textbf{0.589}} & 11.417 & \multicolumn{1}{c|}{\textbf{0.0278}} & \multicolumn{1}{c|}{\textbf{0.4664}} & \multicolumn{1}{c|}{0.526} & \multicolumn{1}{c|}{0.572} & 36.84x \\ \hline
testcase10 & \multicolumn{1}{c|}{1.9483} & \multicolumn{1}{c|}{4.5327} & \multicolumn{1}{c|}{0.086} & 17.53 & \multicolumn{1}{c|}{0.2426} & \multicolumn{1}{c|}{1.2721} & \multicolumn{1}{c|}{0} & \textbf{0.262} & \multicolumn{1}{c|}{0.0659} & \multicolumn{1}{c|}{0.8547} & \multicolumn{1}{c|}{\textbf{0.532}} & 11.270 & \multicolumn{1}{c|}{\textbf{0.0459}} & \multicolumn{1}{c|}{\textbf{0.5741}} & \multicolumn{1}{c|}{0.464} & \multicolumn{1}{c|}{0.551} & 31.81x \\ \hline
testcase13 & \multicolumn{1}{c|}{2.2638} & \multicolumn{1}{c|}{10.5650} & \multicolumn{1}{c|}{0.080} & 3.22 & \multicolumn{1}{c|}{0.2441} & \multicolumn{1}{c|}{3.8389} & \multicolumn{1}{c|}{0} & \textbf{0.033} & \multicolumn{1}{c|}{0.2068} & \multicolumn{1}{c|}{7.2341} & \multicolumn{1}{c|}{0} & 5.452 & \multicolumn{1}{c|}{\textbf{0.0774}} & \multicolumn{1}{c|}{\textbf{2.1299}} & \multicolumn{1}{c|}{\textbf{0.680}} & \multicolumn{1}{c|}{0.204} & 15.78x \\ \hline
testcase14 & \multicolumn{1}{c|}{3.2298} & \multicolumn{1}{c|}{13.1495} & \multicolumn{1}{c|}{0.088} & 2.81 & \multicolumn{1}{c|}{0.3138} & \multicolumn{1}{c|}{5.5321} & \multicolumn{1}{c|}{0} & \textbf{0.033} & \multicolumn{1}{c|}{0.4215} & \multicolumn{1}{c|}{8.8334} & \multicolumn{1}{c|}{0} & 5.463 & \multicolumn{1}{c|}{\textbf{0.1895}} & \multicolumn{1}{c|}{\textbf{2.1035}} & \multicolumn{1}{c|}{\textbf{0.678}} & \multicolumn{1}{c|}{0.203} & 13.84x \\ \hline
testcase15 & \multicolumn{1}{c|}{2.7686} & \multicolumn{1}{c|}{5.7812} & \multicolumn{1}{c|}{0.067} & 8.86 & \multicolumn{1}{c|}{0.1530} & \multicolumn{1}{c|}{1.7691} & \multicolumn{1}{c|}{0} & \textbf{0.105} & \multicolumn{1}{c|}{0.0968} & \multicolumn{1}{c|}{1.5265} & \multicolumn{1}{c|}{0.088} & 8.137 & \multicolumn{1}{c|}{\textbf{0.0353}} & \multicolumn{1}{c|}{\textbf{0.6735}} & \multicolumn{1}{c|}{\textbf{0.733}} & \multicolumn{1}{c|}{0.327} & 27.09x \\ \hline
testcase16 & \multicolumn{1}{c|}{4.6521} & \multicolumn{1}{c|}{7.5669} & \multicolumn{1}{c|}{0.345} & 8.47 & \multicolumn{1}{c|}{0.2675} & \multicolumn{1}{c|}{1.6696} & \multicolumn{1}{c|}{0.258} & \textbf{0.102} & \multicolumn{1}{c|}{0.1601} & \multicolumn{1}{c|}{1.5487} & \multicolumn{1}{c|}{0.529} & 7.413 & \multicolumn{1}{c|}{\textbf{0.0763}} & \multicolumn{1}{c|}{\textbf{0.8393}} & \multicolumn{1}{c|}{\textbf{0.785}} & \multicolumn{1}{c|}{0.324} & 26.14x \\ \hline
testcase19 & \multicolumn{1}{c|}{0.4442} & \multicolumn{1}{c|}{1.7226} & \multicolumn{1}{c|}{0.057} & 23.33 & \multicolumn{1}{c|}{1.0649} & \multicolumn{1}{c|}{1.4689} & \multicolumn{1}{c|}{0} & \textbf{0.281} & \multicolumn{1}{c|}{0.0905} & \multicolumn{1}{c|}{0.5148} & \multicolumn{1}{c|}{0.501} & 11.905 & \multicolumn{1}{c|}{\textbf{0.0187}} & \multicolumn{1}{c|}{\textbf{0.3187}} & \multicolumn{1}{c|}{\textbf{0.752}} & \multicolumn{1}{c|}{0.596} & 39.14x \\ \hline
testcase20 & \multicolumn{1}{c|}{0.6994} & \multicolumn{1}{c|}{2.4261} & \multicolumn{1}{c|}{0.010} & 18.91 & \multicolumn{1}{c|}{0.8204} & \multicolumn{1}{c|}{1.4209} & \multicolumn{1}{c|}{0} & \textbf{0.279} & \multicolumn{1}{c|}{0.1180} & \multicolumn{1}{c|}{0.5003} & \multicolumn{1}{c|}{0.711} & 11.758 & \multicolumn{1}{c|}{\textbf{0.0191}} & \multicolumn{1}{c|}{\textbf{0.3026}} & \multicolumn{1}{c|}{\textbf{0.773}} & \multicolumn{1}{c|}{0.576} & 32.83x \\ \hline \hline
Average & \multicolumn{1}{c|}{1.9785} & \multicolumn{1}{c|}{5.8741} & \multicolumn{1}{c|}{0.168} & 12.2 & \multicolumn{1}{c|}{0.4566} & \multicolumn{1}{c|}{2.1746} & \multicolumn{1}{c|}{0.082} & \textbf{0.166} & \multicolumn{1}{c|}{0.1347} & \multicolumn{1}{c|}{2.4141} & \multicolumn{1}{c|}{0.455} & 8.921 & \multicolumn{1}{c|}{\textbf{0.0533}} & \multicolumn{1}{c|}{\textbf{0.8275}} & \multicolumn{1}{c|}{\textbf{0.723}} & \multicolumn{1}{c|}{0.409} & 27.20x \\ \hline
\end{tabular}

}
\end{table*}

\textbf{\begin{figure*}[t!]
    \centering
    \subfloat[]{{\includegraphics[width=0.6\textwidth]{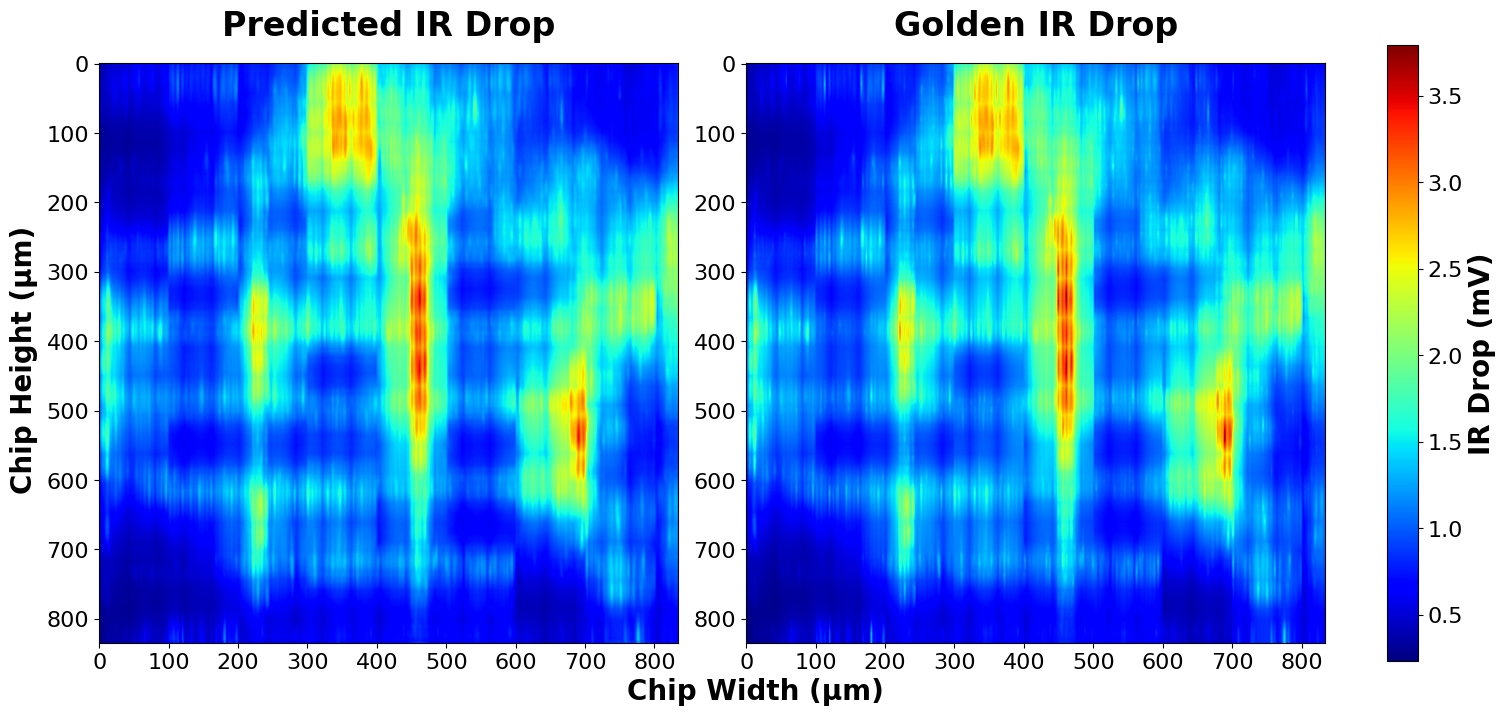} }}
    \qquad
    \subfloat[]{{\includegraphics[width=0.285\textwidth]{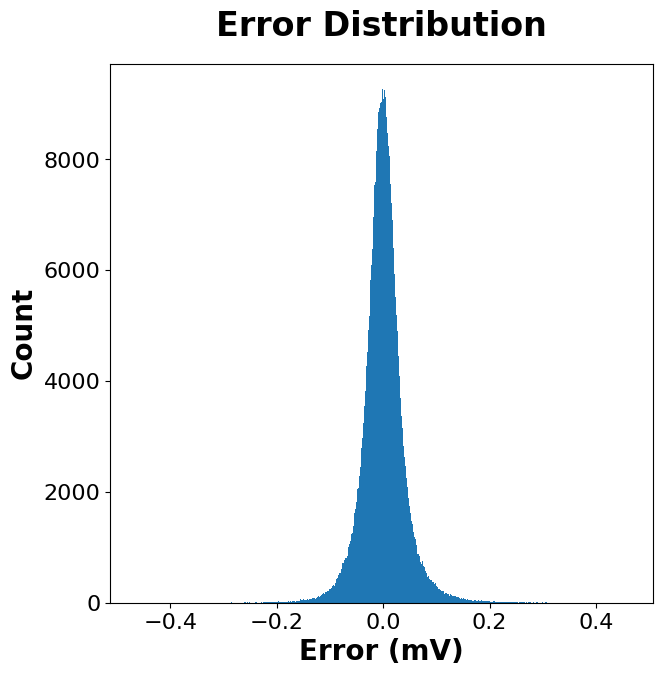} }}
    \caption{(a) CFIRSTNET prediction and the golden IR drop map of testcase 9. (b) Error distribution of benchmark testing.}
    \label{fig:3maps}
\end{figure*}}

\section{Experiment}
\label{sec: exp}

\subsection{Experiment Setup}


CFIRSTNET is implemented with C++, Python3, and PyTorch. All evaluation is conducted on an Intel Xeon E3-1230 v6 CPU server with an NVIDIA Quadro P4000 graphic card. The implementation details can be found at https://github.com/jason122490/CFIRSTNET.

During the training process, a cached dataset is involved in the framework to avoid redundant feature map generation stages. Our selected loss functions are the root-mean-square error (RMSE) and the Dice Loss\cite{Sudre_2017}. This technique guides the model to minimize the error and find the hotspot accurately. Ultimately, the model with the lowest validation loss is saved for testing. On the other hand, all features are resized or generated in the size of 384 x 384 pixels, and z-score normalization is applied to the input features and ground truth.  


The performance of CFIRSTNET is evaluated by an open-source IR drop estimation benchmark\cite{kadagala20232023}. The dataset consists of both irregular PDNs and regular PDNs, including 6100 synthetic data points from BeGAN\cite{chhabria2021began} and 20 real circuits from the OpenROAD project\cite{ajayi2019openroad}. Each of them comes with a current map, a PDN density map, an effective distance map, an IR drop map, and a PDN netlist. Furthermore, the dataset is split into a training set, a validation set, and a testing set. To ensure the feasibility of CFIRSTNET on real circuits, synthetic circuits are assigned to the training set, and the real circuits are distributed to the validation set and testing set equally to match the benchmark condition\cite{kadagala20232023}. The circuits in this dataset are synthesized with open-source NanGate 45nm technology\cite{ajayi2019openroad}. The detail of each test case is provided in Table \ref{tab: circuit}. CFIRSTNET is tested by circuits of various chip areas and PDN planning. Additionally, some of them even involve macros.

Moreover, it is important to mention that the scaling of the IR drop map must be chosen carefully. Though Chhabria et al.\cite{chhabria2021thermal} claimed that higher resolution is inefficient for accuracy, the disproportionate area-to-pixel ratio might not indicate the accurate worst-case IR drop. As exhibited in Figure \ref{fig: resolution}, if we set the resolution too low, we are unable to identify the exact worst-case IR drop nor the precise IR drop value from the downscaled IR drop map. With a more coarse IR drop map, even if significant performance could be achieved, the estimated IR drop values might be far from those in the original circuit. Thus, the circuit resolution (1$\mu$m x 1$\mu$m per pixel) in the open-source benchmark dataset\cite{kadagala20232023} is more reasonable for practical IR drop estimation compared with previous works. As a result, we suggest the resolution used in the evaluation process of CFIRSTNET be identical to the benchmark.

\subsection{Evaluation Metrics}

To assess the performance of CFIRSTNET in comparison to other approaches, we utilized four key evaluation metrics: mean absolute error (MAE), max error, F1 score, and runtime. These metrics align with those reported in the work by Kadagala et al.\cite{kadagala20232023}. 

MAE represents the discrepancy between the predicted IR drop and the actual value, calculated as the sum of absolute errors (Manhattan distance) divided by the sample size Equation (\ref{eq: mae}).

\begin{equation}
    MAE = \dfrac{\sum_{i=1}^{N}|x_i-y_i|}{N}
    \label{eq: mae}
\end{equation}

The F1 score provides a comprehensive view of a model’s ability to correctly classify positive instances while minimizing false positives and false negatives. It combines both precision and recall into a single value, providing a balanced assessment of a model’s performance. In the IR drop estimation task, F1 score could evaluate the model's ability to capture the IR drop hot spot, i.e., the region with IR drop larger than 90\% of the testcase, according to Kadagala et al.\cite{kadagala20232023}. As shown in Equation (\ref{eq: f1}), the F1 score is obtained with true positive (TP), false positive (FP), and false negative (FN) regions.

\begin{equation}
    F1 Score = \frac{2*TP}{2*TP + FP + FN}
    \label{eq: f1}
\end{equation}

Moreover, the runtime is also measured in order to show whether CFIRSTNET could speed up the PDN analysis time. It is imperative to mention that all experiments are carried out on the same machine to make fair comparisons.

\section{Evaluation Result}
\label{sec: result}

\subsection{Benchmark Performance}

Figure \ref{fig:3maps}(a) is an example of the predicted IR drop map and the golden one. According to Figure \ref{fig:3maps}(b), which is the error distribution of benchmark testing, the errors of the majority of predicted IR drops are within 0.2mV. Numerically, Table \ref{tab: result-table} presents the MAE $e_{avg}$, maximum absolute error, $e_{max}$, F1 score, and runtime of each approach. HSPICE is selected as the golden IR drop tool. CFIRSTNET is compared with IREDGe from \cite{chhabria2021thermal} and 1st Place of CAD Contest from \cite{kadagala20232023}. IREDGe\footnote{IREDGe method is reproduced with its source code from GitHub despite the discrepancy between the code and the paper.} can make IR drop estimations instantly but the $e_{avg}$ and $e_{max}$ is not negligible. Though the 1st place winner in the ICCAD CAD Contest 2023\cite{kadagala20232023} made good estimations, the $e_{avg}$ is still improvable and the hotspot in some testcases could not be located. CFIRSTNET achieves a 13-39x speedup compared to HSPICE, $e_{avg}$ lower than 0.19mV, $e_{max}$ lower than 2.13mV. Given that the IR drop constraint normally falls into 1\% to 2.5\% of VDD (1.1V in this case), the result indicates that the estimations made by CFIRSTNET could provide lower $e_{avg}$ with reasonable $e_{max}$ and meanwhile ensure efficiency. Moreover, owing to the properties of ML methods, the runtime of CFIRSTNET does not exhibit the same increase as HSPICE does with respect to the growth of chip size.

\subsection{Compared with CAD Contest Results}

\begin{table}[h!]
\centering
\caption{Compared with winners of ICCAD CAD Contest 2023}
\smallskip
\label{tab: cad_com}
\resizebox{1\linewidth}{!}{
\begin{tabular}{@{}cccc@{}}
\toprule
                             & $e_{avg}$ (mV)  & F1 score       & runtime (sec)  \\ \midrule
    1st place winner         & 0.1347          & 0.455          & 8.921         \\
    2nd place winner         & 0.1498          & 0.455          & N/A          \\
    \textbf{CFIRSTNET(Ours)} & \textbf{0.0533} & \textbf{0.723} & \textbf{0.409} \\ 
\bottomrule
\end{tabular}
}
\end{table}

Table \ref{tab: cad_com} shows the evaluation result of CFIRSTNET compared with the hitherto best solutions presented in the CAD Contest\footnote{We have obtained the 1st Place's executable but that of 2nd Place is not available.}\cite{kadagala20232023}. By virtue of the comprehensive features and tailored model, CFIRSTNET reduces $e_{avg}$ by 60\%, increases F1 score by 59\%, and achieves a 21.8x speedup compared to 1st Place winner. Moreover, though the CAD Contest didn't evaluate $e_{max}$, CFIRSTNET reduces $e_{max}$ by 66\% (Table \ref{tab: result-table}). After making a direct comparison under identical benchmark conditions, we could state that CFIRSTNET is currently the state-of-the-art solution to the static IR drop estimation problems.

\subsection{Ablation Studies of Comprehensive Features}

\begin{table}[h!]
\centering
\caption{Performance with features involved gradually. }
\smallskip
\label{tab: ablation-table}
\resizebox{1\linewidth}{!}{
\begin{tabular}{@{}cccccc@{}}
\toprule
                             & $e_{avg}$  & $e_{max}$  & F1 score   & runtime    \\ \midrule
Image-based                  & 0.1864     & 2.2650     & 0.217      & \textbf{0.290}        \\
HIRD                         & 0.0795     & 0.9341     & 0.677      & 0.347        \\
HIRD+WR+RD                   & 0.0612     & 0.8691     & 0.682      & 0.356        \\
All Features                 & 0.0559     & 0.9351     & 0.674      & 0.486        \\
\textbf{Comprehensive Features}       & \textbf{0.0533}     & \textbf{0.8275}     & \textbf{0.723}      & 0.409        \\
\bottomrule
\end{tabular}
}
\end{table}
 
Table \ref{tab: ablation-table} compares the effectiveness of the proposed features gradually. The image-based input established a baseline performance. The proposed Hypothetical IR Drop Distillation (HIRD) boosted the experimental result substantially. With the Wire Resistance (WR) maps and the Resistive Distance (RD) maps coming into practice, $e_{avg}$ and $e_{max}$ become lower and a higher F1 score is achieved. When all features are taken, CFIRSTNET achieves significant improvement compared to the baseline.

\begin{figure}[h!]
    \includegraphics[width=\linewidth]{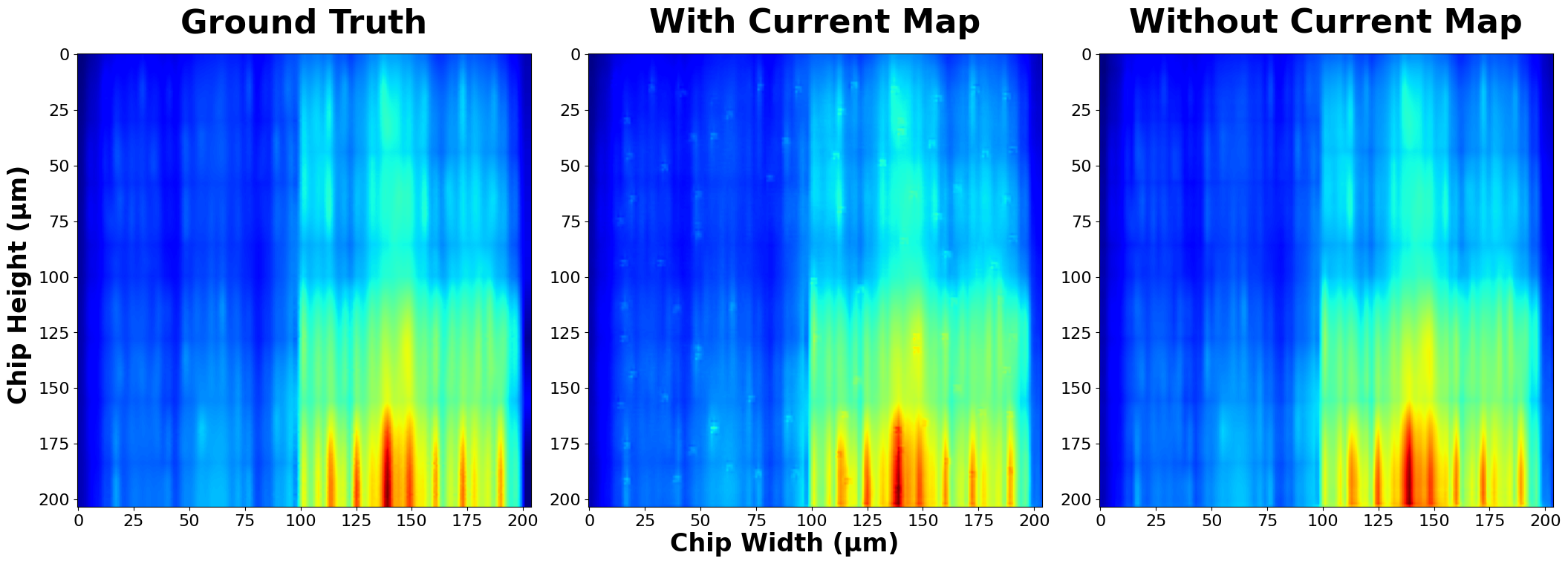}
    \caption{IR drop estimation maps with or without involving current map.}
    \label{fig: current map}
\end{figure}

Moreover, refer to Figure \ref{fig: current map}, we noticed that incorporating the current map could cause some abnormal spikes in terms of IR drop estimation. Thus, since the Hypothetical IR Drop Distillation has incorporated information on the cell-level current, the Comprehensive features would involve all features but the current map. As a result, the performance was improved slightly without the current map as input.

\section{Conclusion}
\label{sec: conclusion}

In this paper, we present CFIRSTNET, a novel approach for static IR drop estimation. CFIRSTNET includes comprehensive feature extraction and a custom-designed CNN. Consequently, CFIRSTNET could make IR drop estimations within 0.6 seconds and achieve a 13-39x speedup compared to HSPICE with $e_{avg}$ below 0.19mV and maximum error less than 2.13mV. When evaluated with the open-source ICCAD CAD Contest 2023 benchmark, CFIRSTNET outperformed previous works in terms of MAE, max error, F1 score, and runtime. Moreover, CFIRSTNET achieved recognizable results compared to previous works. In terms of the possibilities of future works, more open-source PDNs could be incorporated to further explore the improvability of CFIRSTNET on this problem or even dynamic IR drop estimation.

\begin{acks}
This work was supported by the National Science and Technology Council of Taiwan under Grant 112-2221-E-A49-156-MY3.
\end{acks}
\bibliographystyle{ACM-Reference-Format}
\bibliography{references.bib}

\end{document}